# PJAIT Systems for the IWSLT 2015 Evaluation Campaign Enhanced by Comparable Corpora

*Krzysztof Wołk, Krzysztof Marasek*

Multimedia Department
Polish-Japanese Academy of Information Technology, Koszykowa 86, 02-008 Warsaw
kwolk@pja.edu.pl, kmarasek@pja.edu.pl

**Abstract**

In this paper, we attempt to improve Statistical Machine Translation (SMT) systems on a very diverse set of language pairs (in both directions): Czech - English, Vietnamese - English, French - English and German - English. To accomplish this, we performed translation model training, created adaptations of training settings for each language pair, and obtained comparable corpora for our SMT systems. Innovative tools and data adaptation techniques were employed. The TED parallel text corpora for the IWSLT 2015 evaluation campaign were used to train language models, and to develop, tune, and test the system. In addition, we prepared Wikipedia-based comparable corpora for use with our SMT system. This data was specified as permissible for the IWSLT 2015 evaluation. We explored the use of domain adaptation techniques, symmetrized word alignment models, the unsupervised transliteration models and the KenLM language modeling tool. To evaluate the effects of different preparations on translation results, we conducted experiments and used the BLEU, NIST and TER metrics. Our results indicate that our approach produced a positive impact on SMT quality.

## 1. Introduction

Statistical Machine Translation (SMT) must deal with a number of problems to achieve high quality. These problems include the need to align parallel texts in language pairs and cleaning harvested parallel corpora to remove errors. This is especially true for real-world corpora developed from text harvested from the vast data available on the Internet. Out-Of-Vocabulary (OOV) words must also be handled, as they are inevitable in real-world texts [1].

The lack of enough parallel corpora is another significant challenge for SMT. Since the approach is statistical in nature, a significant amount of quality language pair data is needed to improve translation accuracy. In addition, very general translation systems that work in a general text domain have accuracy problems in specific domains. SMT systems are more accurate on corpora from a domain that is not too wide. This exacerbates the data problem, calling for the enhancement of comparable corpora for particular text domains [2].

This paper describes SMT research that addresses these problems, particularly comparable corpora within the limits of permissible data for the IWSLT 2015 campaign. We selected a diverse set of language pairs for translation in both directions: Czech and English, Vietnamese and English, French and English, and German and English. To accomplish this, we performed model training, created adaptations of training settings and data for each language pair, and enhanced our systems by building and using comparable corpora in our statistical translation systems.

Innovative tools and data adaptation techniques were employed. The Technology, Entertainment and Design (TED) parallel text corpora for the IWSLT 2015 evaluation campaign were used to train language models, and to develop, tune, and test the system. In addition, we prepared Wikipedia-based comparable corpora for use with our SMT systems. We explored the use of domain adaptation techniques, symmetrized word alignment models, the unsupervised transliteration models, and the KenLM language modeling tool [3]. To evaluate the effects of different preparations on translation results, we conducted experiments and evaluated the results using standard SMT metrics [4].

The languages translated during this research are diverse: Czech, English, French, German, and Vietnamese. The first four belong to three different branches of the Indo-European language family. Czech is found in the Slavic branch of that language family. English and German fall in the Western group of the Germanic branch of the Indo-European family, while French is found in the Romance branch. Vietnamese falls into an entirely different language family, Austro-Asiatic. So, our translation challenges cross languages, language branches, and language families [5, 6, 7, 8, 9].

This paper is structured as follows: Section 2 explains the data preparation. Section 3 presents experiment setup and the results. Lastly in Section 4 we summarize the work.

## 2. Data preparation

This section describes our techniques for data preparation for our SMT systems. We give particular emphasis to preparation of the language data and models and our domain adaptation approach.

### 2.1. Obtaining Comparable Corpora

We provided a new approach for mining parallel corpora from Wikipedia in support of IWSLT 2015 objectives. In general, Wikipedia data is noisy, has a wide domain, and the sentences of its bilingual texts are not aligned. To extract additional data from Wikipedia, we used the method described in [17] and adapted it for the needs of new languages.

Our tool performs model training and phrase-level symmetrization in a multi-threaded approach to increase performance. The dedicated tuning tool is used to determine the trained model's optimal weights [17].

To be more precise our approach enhances the Yalign tool [18], which is single threaded and becomes computationally infeasible for large-scale mining of corpora. Yalign's existing classifier also requires individual pairs of input text or webpages to be loaded into memory for alignment, and the classifier must be reloaded. To increase performance, we modified the design to supply the classifier with Wikipedia articles within one session, with no need to reload the classifier. Performance improvements of more than a factor of 6x have been observed, as a result [17].

Our approach also replaces Yalign's alignment algorithm with a GPU-optimized version of the Needleman-Wunsch algorithm. In addition to the performance improvements this brings, it also yields higher alignment accuracy. A tuning algorithm enables automatic selection of threshold and penalty parameters to adaptably perform tradeoffs between precision and recall [17].

Wikipedia data in the following language pairs were mined using our tool: English-Czech, English-German, English-French, and English-Vietnamese. This dataset was accepted by the IWSLT 2015 evaluation organizers as permissible data [19].

### 2.2. Data Preparation

Five languages were involved in this research: Czech, English, French, German, and Vietnamese. TED talks training data for those languages consisted of the following:

- Czech: approx. 11 MB - 176,094 untokenized words
- French: approx. 24 MB - 165,605 untokenized words
- German: approx. 22 MB - 213,486 untokenized words
- Vietnamese: approx. 18 MB, 52,549 untokenized words
- English: approx. 125,000 untokenized words for DE and FR, 85,000 words for CS and 100,000 for VI pairs

The TED transcripts prepared by the FBK team for IWSLT[1] consist of text encoded in UTF-8 format, separated into sentences, and provided in pairs of languages. The data is provided as XML files [1].

Pre-processing, both automatic and manual, of this training data was required. There were a variety of errors found in this data, including spelling errors, unusual nesting of text, text duplication, and parallel text issues. Approximately 2% of the text in the training set contained spelling errors, and approximately 4% of the text had insertion errors. A tool described in [2], was used to correct these errors. Previous studies have found that such cleaning increases the BLEU score for SMT by a factor of 1.5–2 [1].

After cleaning and tokenization, we found the following amounts of unique word forms in the different languages:
- Czech: 109,692 tokenized words
- French: 76,216 tokenized words
- German: 123,673 tokenized words
- Vietnamese: 24,914 tokenized words
- English: approx. 53,000 tokenized words for DE and FR, 39,000 tokenized words for CS and 44,000 for VI pairs

In addition, comparable corpora in those language pairs were created from Wikipedia data. The Wikipedia data obtained consisted of the following:
- Czech: approx. 22 MB - 104,698 untokenized words in 27,723 parallel sentences
- French: approx. 101 MB – 1,121,424 untokenized words in 818,300 parallel sentences
- German: approx. 277 MB – 2,865,865 untokenized words in 2,459,662 parallel sentences
- Vietnamese: approx. 33 MB - 93,218 untokenized words in 58,116 parallel sentences
- English: approx. 285MB – 2,577,854 untokenized words in DE pair, approx. 113MB – 1,290,499 untokenized words in FR pair, approx. 22MB – 98,820 untokenized words in CS pair and approx. 34MB – 92,445 untokenized words in VI pair

After cleaning [2] and tokenization:
- Czech: approx. 73,996 words in CS side and 65,592 words at EN side
- French: approx. 517,604 words in FR side and 544,994 words at EN side
- German: approx. 1,327,789 words in DE side and 922,785 words at EN side
- Vietnamese: approx. 66,391 words in FR side and 65,852 words at EN side

SyMGiza++, a tool that supports the creation of symmetric word alignment models, was used to extract parallel phrases from the data. This tool enables alignment models that support many-to-one and one-to-many alignments in both directions between two language pairs. SyMGiza++ is also designed to leverage the power of multiple processors through advanced threading management, making it very fast. Its alignment process uses four different models during training to progressively refine alignment results. This approach has yielded impressive results in [10].

Out-Of-Vocabulary (OOV) words pose another significant challenge to SMT systems. If not addressed, unknown words appear, untranslated, in the output, lowering the translation quality. To address OOV words, we used implemented in the Moses toolkit Unsupervised Transliteration Model (UTM). UTM is an unsupervised, language-independent approach for learning OOV words. We used the post-decoding transliteration option with this tool. UTM uses a transliteration phrase translation table to evaluate and score multiple possible transliterations [11, 12].

The KenLM tool was applied to the language model to train and binarize it. This library enables highly efficient queries to language models, saving both memory and computation time. The lexical values of phrases are used to condition the reordering probabilities of phrases. We used KenLM with lexical reordering set to hier-msd-bidirectional-fe. This setting uses a hierarchical model that considers three orientation types based on both source and target phrases: monotone (M), swap (S), and discontinuous (D). Probabilities of possible phrase orders are examined by the bidirectional reordering model [3, 13, 14].

### 2.3. Domain Adaptation

The TED data sets have a rather a wide domain, but rather not as wide-ranging in topic as the Wikipedia articles. Since SMT systems work best in a defined domain, this presents another considerable challenge. If not addressed, this would lead to lower translation accuracy.

The quality of domain adaptation depends heavily on training data used to optimize the language and translation models in an SMT system. Selection and extraction of domain-specific training data from a large, general corpus addresses this issue [15]. This process uses a parallel, general domain corpus and a general domain monolingual corpus in the target language. The result is a pseudo in-domain sub-corpus.

---
[1] iwslt.org

As described by Wang et al. in [16], there are generally three processing stages in data selection for domain adaptation. First, sentence pairs from the parallel, general domain corpus are scored for relevance to the target domain. Second, resampling is performed to select the best-scoring sentence pairs to retain in the pseudo in-domain sub-corpus. Those two steps can also be applied to the general domain monolingual corpus to select sentences for use in a language model. After collecting a substantial amount of sentence pairs (for the translation model) or sentences (for the language model), those models are trained on the sub-corpus that represents the target domain [16].

Similarity measurement is required to select sentences for the pseudo in-domain sub-corpus. There are three state-of-the-art approaches for similarity measurement. The cosine tf-idf criterion looks for word overlap in determining similarity. This technique is specifically helpful in reducing the number of OOV words, but it is sensitive to noise in the data. A perplexity-based criterion considers the $n$-gram word order in addition to collocation. Lastly, edit distance simultaneously considers word order, position, and overlap. It is the strictest of the three approaches. In their study [16], Wang et al. found that a combination of these approaches provided the best performance in domain adaptation for Chinese-English corpora [16].

In accordance with Wang et al.'s approach [16], we use a combination of the criteria at both the corpora and language models. The three similarity metrics are used to select different pseudo in-domain sub-corpora. The sub-corpora are then joined during resampling based on a combination of the three metrics. Similarly, the three metrics are combined for domain adaptation during translation. We empirically found acceptance rates that allowed us only to harvest 20% of most domain-similar data [16].

## 3. Experimental Results

Various versions of our SMT systems were evaluated via experimentation. In preparation for experiments, we processed the corpora. This involved tokenization, cleaning, factorization, conversion to lower case, splitting, and final cleaning after splitting. Language models were developed and tuned using the training data.

The Experiment Management System [4] from the open source Moses SMT toolkit was used to conduct the experiments. Training of a 6-gram language model was accomplished our resulting systems using the KenLM Modeling Toolkit instead of 5-gram SRILM [20] with an interpolated version of Kneser-Key discounting (interpolate – unk –kndiscount) that was used in our baseline systems. Word and phrase alignment was performed using SyMGIZA++ [10] instead of GIZA++. KenLM was also used, as described earlier, to binarize the language models. The OOV's were handled by using Unsupervised Transliteration Model [12].

The results are shown in Table 1 and 2. Each language pair was translated in both directions. "BASE" in the tables represents the baseline SMT system. "EXT" indicates results for the baseline system, using the baseline settings but extended with comparable corpora from Wikipedia. "BEST" indicates the results when the new SMT settings were applied and using all permissible data. For DE and FR we did not train systems using more permissible data that our Wikipedia comparable corpora. Additionally, we conducted progressive tests only for FR and DE data because CS and VI were not evaluated before during IWSLT campaigns.

Three well-known metrics were used for scoring the results: Bilingual Evaluation Understudy (BLEU), the US National Institute of Standards and Technology (NIST) metric and Translation Error Rate (TER).

In addition to TED data, the data permissible for the IWSLT 2015 campaign included: data from the Workshop on Machine Translation (WMT) 2015 web page [21], MultiUN data [22, 23] (translated United Nations documents) and parallel corpora we provided from the Wikipedia [19].

The results show that the systems extended with comparable corpora from Wikipedia performed well on all data sets in comparison to the baseline SMT systems. Application of the new settings and use of all permissible data improved performance even more.

*Table 1:* Progressive Results, 2014 Test Data

| LANG | SYSTEM | DIRECTION | BLEU | NIST | TER |
|---|---|---|---|---|---|
| DE-EN | BASE | →EN | 17.99 | 5.51 | 64.35 |
|  | EXT | →EN | 21.92 | 6.04 | 60.58 |
|  | BASE | ←EN | 18.49 | 5.74 | 61.65 |
|  | EXT | ←EN | 20.68 | 5.99 | 59.77 |
| FR-EN | BASE | →EN | 32.20 | 7.36 | 47.60 |
|  | EXT | →EN | 32.92 | 7.37 | 48.25 |
|  | BASE | ←EN | 30.31 | 7.24 | 50.17 |
|  | EXT | ←EN | 31.88 | 7.49 | 47.92 |

*Table 2:* Results, 2015 Test Data

| LANG | SYSTEM | DIRECTION | BLEU | NIST | TER |
|---|---|---|---|---|---|
| DE-EN | BASE | →EN | 21.78 | 6.49 | 55.45 |
|  | EXT | →EN | 26.08 | 7.03 | 54.34 |
|  | BASE | ←EN | 20.08 | 5.76 | 61.37 |
|  | EXT | ←EN | 22.51 | 6.04 | 59.02 |
| FR-EN | BASE | →EN | 31.94 | 7.34 | 47.55 |
|  | EXT | →EN | 32.75 | 7.27 | 48.40 |
|  | BASE | ←EN | 30.54 | 6.99 | 51.51 |
|  | EXT | ←EN | 32.79 | 7.32 | 49.15 |
| CS-EN | BASE | →EN | 22.44 | 6.11 | 57.98 |
|  | EXT | →EN | 24.19 | 6.03 | 56.13 |
|  | BEST | →EN | 25.07 | 6.40 | 55.74 |
|  | BASE | ←EN | 14.74 | 4.74 | 65.80 |
|  | EXT | ←EN | 15.18 | 4.86 | 65.11 |
|  | BEST | ←EN | 17.17 | 5.10 | 63.00 |
| VI-EN | BASE | →EN | 24.61 | 5.92 | 59.32 |
|  | EXT | →EN | 22.41 | 5.68 | 63.78 |
|  | BEST | →EN | 23.46 | 5.73 | 62.19 |
|  | BASE | ←EN | 27.01 | 6.47 | 58.42 |
|  | EXT | ←EN | 27.16 | 6.23 | 66.18 |
|  | BEST | ←EN | 28.39 | 6.67 | 65.01 |

## 4. Summary

We have improved SMT for a very diverse set of language pairs, in both translation directions, using data permissible for the IWSLT 2015 evaluation campaign. We cleaned, prepared, and tokenized the training data. Symmetric word alignment models were used to align the corpora. UTM was used to handle OOV words. A language model was created, binarized, and tuned. We performed domain adaptation of language data using a combination of similarity metrics.

Experiments were performed using the data permissible by the IWSLT 2015 organizers. The results show a positive impact of our approach on SMT quality across the language pairs. Only surprising result was in translation from Vietnamese into English, where our best system outperformed the baseline. We conducted detailed research regarding this issue, including tuning results for each iteration and evaluation of each TED talk separately. We found out that on most talks our system worked correctly only two of them the results were negative. The talk number 2183 the baseline BLEU score was equal to 63.88 (BASE) whereas our system score (BEST) was equal to 49.73. Such big disproportion is most likely reason for strange overall evaluation score. We believe that some parts of talk 2183 were present in training data, and extending this data decreased the scores. Detailed evaluation results are presented in the Table 3. Additionally, we can suspect, from the statistics presented in Chapter 2.2, that Wikipedia data for Vietnamese is was not good enough. Having 93,218 words in 58,116 sentences would mean that this corpus basically consists of uni- or bi-grams.

*Table 3: Detailed Vietnamese-English Results*

| TALK ID | SYSTEM | BLEU |
|---|---|---|
| 1922 | BASE | 17.25 |
|      | BEST | 18.07 |
| 1932 | BASE | 15.70 |
|      | BEST | 18.17 |
| 1939 | BASE | 12.35 |
|      | BEST | 13.26 |
| 1954 | BASE | 27.96 |
|      | BEST | 28.92 |
| 1961 | BASE | 19.59 |
|      | BEST | 21.35 |
| 1997 | BASE | 16.81 |
|      | BEST | 19.67 |
| 2007 | BASE | 16.38 |
|      | BEST | 19.67 |
| 2017 | BASE | 21.08 |
|      | BEST | 22.42 |
| 2024 | BASE | 9.44 |
|      | BEST | 6.25 |
| 2045 | BASE | 21.05 |
|      | BEST | 21.09 |
| 2102 | BASE | 17.14 |
|      | BEST | 20.16 |
| 2183 | BASE | 63.88 |
|      | BEST | 49.73 |